\documentclass{entcs} \usepackage{prentcsmacro}

\usepackage[english]{babel}
\usepackage{enumerate}
\usepackage[shortlabels]{enumitem}
\usepackage{amsmath}
\usepackage{amssymb}
\usepackage{mathtools}
\usepackage{graphicx,epsfig,color}
\usepackage[color]{xypic}
\input xy
\xyoption{all}
\xyoption{2cell}
\UseAllTwocells
\usepackage{comment}
\usepackage{etoolbox}
\usepackage{multicol}
\usepackage{mathrsfs} 

\usepackage{longtable}
\usepackage{wrapfig}
\usepackage[table]{xcolor}
\usepackage{multirow}

\usepackage{array}
\usepackage{physics}
\usepackage{fancybox}
\usepackage{nicefrac}

\newenvironment{myproof}{\noindent\textbf{Proof}}{}
\newcommand{\QEDbox}{\square}
\newcommand{\QED}{\hspace*{\fill}$\QEDbox$}

\def \after {\mathrel{\circ}}

\def \N {\mathbb{N}}
\def \R {\mathbb{R}}
\newcommand{\leftScottint}{[{\kern-.3ex}[}
\newcommand{\rightScottint}{]{\kern-.3ex}]}
\newcommand{\Scottint}[1]{\leftScottint\,#1\,\rightScottint}

\newcommand{\set}[2]{\{#1\;|\;#2\}}
\newcommand{\setin}[3]{\{#1\in#2\;|\;#3\}}

\newcommand{\tuple}[1]{\langle#1\rangle}

\newcommand\Mlt{\mathcal{M}}
\newcommand\Pow{\mathcal{P}}

\def \Set {\mathbf{Set}}
\def \NN  {\mathbf{NN}} 
\def \RF  {\mathbf{RF}} 
\def \SL  {\mathbf{SL}} 
\renewcommand{\op}[1]{#1^{\mathrm{op}}}
\def \Hom {\mathrm{Hom}}

\def \Stat {\mathrm{Stat}}
\def \Pred {\mathrm{Pred}}
\newcommand{\bigket}[1]{\ensuremath{\big|{\kern.1em}#1{\kern.1em}\big\rangle}}
\newcommand{\supp}{\mathrm{supp}}
\newcommand{\Kl}[1]{\mathcal{K}{\kern-.4ex}\ell(#1)}
\newcommand{\intd}{{\kern.2em}\mathrm{d}{\kern.03em}}

\def\lastname{Jacobs and Sprunger}

\begin{document}

\begin{frontmatter}

\title{Neural Nets via \\ Forward State Transformation and \\ Backward Loss
  Transformation}

\author{Bart Jacobs\thanksref{myemail}} 
\qquad 
\author{David Sprunger\thanksref{coemail}}

\address{Institute for Computing and Information Sciences (iCIS) \\ 
   Radboud University Nijmegen \\ 
   The Netherlands}
\address{ERATO Metamathematics for Systems Design Project \\ 
   National Insitute of Informatics \\ 
   Tokyo, Japan}

\thanks[myemail]{Email:\href{mailto:bart@cs.ru.nl}
   {\texttt{\normalshape bart@cs.ru.nl}}} 
\thanks[coemail]{Email:\href{mailto:sprunger@nii.ac.jp} 
   {\texttt{\normalshape sprunger@nii.ac.jp}}}

\begin{abstract}
This article studies (multilayer perceptron) neural networks with an
emphasis on the transformations involved --- both forward and backward
--- in order to develop a semantical/logical perspective that is in
line with standard program semantics.  The common two-pass neural
network training algorithms make this viewpoint particularly
fitting. In the forward direction, neural networks act as state
transformers. In the reverse direction, however, neural networks
change losses of outputs to losses of inputs, thereby acting like a
(real-valued) predicate transformer. In this way, backpropagation is
functorial by construction, as shown earlier in recent other work. We
illustrate this perspective by training a simple instance of a neural
network.
\end{abstract}

\begin{keyword}
Neural network, backpropagation, multilayer perceptron, state-and-effect
triangle, loss transformation
\end{keyword}

\end{frontmatter}

\section{Introduction}\label{sec:intro}

Though interest in artificial intelligence and machine learning have
always been high, the public's exposure to successful applications has
markedly increased in recent years. From consumer-oriented applications
like recommendation engines, speech face recognition, and text
prediction to prominent examples of superhuman performance (DeepMind's
AlphaGo, IBM's Watson), the impressive results of machine learning
continue to grow.

Though the understandable excitement around the expanding catalog of
successful applications lends a kind of mystique, neural networks and
the algorithms which train them are, at their core, a special kind of
computer program. One perspective on programs which is relevant in this
domain are so-called \emph{state-and-effect triangles}, which emphasize
the dual nature of programs as both state and predicate transformers.
This framework originated in quantum computing, but has a wide variety
of applications including deterministic and probabilistic
computations~\cite{Jacobs17b}.

The common two-pass training scheme in neural networks makes their
dual role particularly evident. Operating in the ``forward direction''
neural networks are like a function: given an input signal they behave
like (a mathematical model of) a brain to produce an output
signal. This is a form of state transformation. In the ``backwards
direction'', however, the derivative of a loss function with respect
to the output of the network is
\emph{backpropagated}~\cite{Rumelhart86} to the derivative of the loss
function with respect to the inputs to the network. This is a kind of
predicate transformation, taking a real-valued predicate about the
loss at the output and producing a real-valued predicate about the
source of loss at the input. The main novel perspective offered by
this paper uses such state-and-effect `triangles' for neural
networks. We expect that such more formal approaches to neural
networks can be of use in trends towards \emph{explainable} AI, where
the goal is to extend automated decisions/classifications with human
understandable explanations.

In recent years, it has become apparent that the architecture of a
neural network is very important for its accuracy and trainability in
particular problem domains~\cite{Goodfellow16}. This has resulted in a
profligation of specialized architectures, each adapted to its
application. Our goal here is not to express the wide variety of
special neural networks in a single framework, but rather to describe
neural networks generally as an instance of this duality between state
and predicate transformers. Therefore, we shall work with a simple,
suitably generic neural network type called the \emph{multilayer
  perceptron} (MLP).

We see this paper as one of recent steps towards the application of
modern semantical and logical techniques to neural networks, following
for instance~\cite{FongST17,GhicaMCDR18}.

\emph{Outline.} In this paper, we begin by describing MLPs, the layers
they are composed of, and their forward semantics as a state
transformation (Section 2). In Section 3, we give the corresponding
backwards transformation on loss functions and use that to formulate
backpropagation in Section 4. Finally, in Section 5, we discuss the
compositional nature of backpropagation by casting it as a functor,
and compare our work in particular to~\cite{FongST17}.

\section{Forward state transformation}

Much like ordinary programs, neural networks are often subdivided into
functional units which can then be composed both in sequence and in
parallel. These subnetworks are usually called \emph{layers}, and the
sequential composition of several layers is by definition a ``deep''
network\footnote{In contrast, the ``width'' of a layer typically
  refers to the number of input and output units, which can be thought
  of as the repeated parallel composition of yet another
  architecture.}.  There are a number of common layer types, and a
neural network can often be described by naming the layer types and
the way these layers are composed.

\emph{Feedforward networks} are an important class of neural networks
where the composition structure of layers forms a directed acyclic
graph---the layers can be put in an order so that no layer is used as
the input to an earlier layer. A \emph{multilayer perceptron} is a
particular kind of feedforward network where all layers have the same
general architecture, called a fully-connected layer, and are composed
strictly in sequence. As mentioned in the introduction, the MLP is
perhaps the prototypical neural network architecture, so we treat this
network type as a representative example. In the sequel, we will use
the phrase ``neural network'' to denote this particular network
architecture.

More concretely, a layer consists of two lists of nodes with directed
edges between them. For instance, a neural network with two layers may
be depicted as follows.
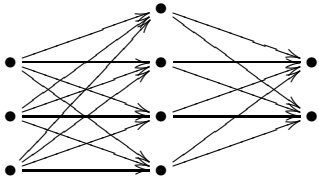
\[ \xymatrix@R-1pc@C+2pc{
& \bullet\ar[dr]\ar[ddr] &
\\
\bullet\ar[ur]\ar[r]\ar[dr]\ar[ddr] 
   & \bullet\ar[r]\ar[dr]
   & \bullet
\\
\bullet\ar[uur]\ar[ur]\ar[r]\ar[dr]
   & \bullet\ar[ur]\ar[r]
   & \bullet
\\
\bullet\ar[uuur]\ar[uur]\ar[ur]\ar[r]
   & \bullet\ar[uur]\ar[ur]
} \]

\noindent We will represent such a network via special arrows $3
\Rightarrow 4 \Rightarrow 2$, where the numbers 3, 4, and 2 correspond
to the number of nodes at each stage. These arrows involve weights,
biases, masks, and activations, see Definition~\ref{def:layer} below.
The (forward) semantics of these arrows is given by functions $\R^{3}
\rightarrow \R^{4} \rightarrow \R^{2}$. They will be described in
greater detail shortly, in Definition~\ref{def:forward}. We first
concentrate on individual layers.

In the definition below we shall write $\Mlt(n) = \R^n$ and $\Pow(n) =
\setin{k}{\N}{k \subseteq n}$. In this description of the powerset
$\Pow$ we identify a natural number $n\in\N$ with the $n$-element
subset of numbers $\{0,1,\ldots,n-1\}$ below $n$. We shall have more
to say about $\Mlt$ and $\Pow$ in Remark~\ref{rem:monad} below.

\begin{definition}
\label{def:layer}
A single \emph{layer} $n\Rightarrow k$ between natural numbers
$n,k\in\N$ is given by three functions:
\[ \begin{array}{ccl}
\xymatrix{ n+1\ar[r]^-{T} & \Mlt(k) }
& \qquad\qquad &
\mbox{the transition function}
\\
\xymatrix{ n\ar[r]^-{M} & \Pow(k) }
& &
\mbox{the mask function}
\\
\xymatrix{ \R\ar[r]^-{\alpha} & \R }
& &
\mbox{the activation function.}
\end{array} \]

\noindent The transition function $T$ can be decomposed into a pair
$[T_{w}, T_{b}]$, where $T_{w} \colon n \rightarrow \Mlt(k)$ captures
the weights and $T_{b}\in\Mlt(k)$ the biases. The mask function $M
\colon n \rightarrow \Pow(k)$ captures connections and mutability; it
works as follows, for $i\in n$ and $j\in k$.
\[ \left\{\begin{array}{rcl}
j \in M(i)
& \mbox{\quad means \quad} &
\begin{minipage}[t]{13em}
there is a mutable connection from node $i$ to node $j$, with weight
$\Mlt(i)(j)$
\end{minipage}
\\
j\not\in M(i) \mbox{ and } \Mlt(i)(j) = 0
& \mbox{\quad means \quad} &
\begin{minipage}[t]{13em}
there is \emph{no} connection from node $i$ to node $j$
\end{minipage}
\\
j\not\in M(i) \mbox{ and } \Mlt(i)(j) \neq 0
& \mbox{\quad means \quad} &
\begin{minipage}[t]{13em}
there is a \emph{non-mutable} connection from node $i$ to node $j$,
with weight $\Mlt(i)(j)$.
\end{minipage}
\end{array}\right. \]

\noindent The activation function $\alpha\colon\R\rightarrow\R$ is
required to be differentiable. 
\end{definition}

Mutability is used only to determine which weights should be updated
after back propagation. In particular, $M$ is not used in forward
propagation, and we often omit $M$ in situations where it plays no role,
including forward propagation.

\begin{remark}
\label{rem:monad}
The operations $\Mlt$ and $\Pow$ are called \emph{multiset} and
\emph{powerset}. They both form a monad on the category $\Set$ of sets
and functions. In general, they are defined on a set $I$ as:
\[ \begin{array}{rcl}
\Pow(I)
& = &
\set{S}{S\subseteq I}
\\
\Mlt(I)
& = &
\set{\varphi\colon I \rightarrow \R}{\supp(\varphi) \mbox{ is finite}},
\end{array} \]

\noindent where $\supp(\varphi) = \setin{i}{I}{\varphi(i) \neq 0}$ is
the support of $\varphi$. Such a function $\varphi$ can also be written
as formal sum:
\[ \begin{array}{rclcl}
\varphi
& \equiv &
r_{1}\ket{i_{1}} + \cdots + r_{m}\ket{i_m}
& \mbox{\quad where \quad} &
\left\{\begin{array}{l}
\supp(\varphi) = \{i_{1}, \ldots, i_{m}\} \subseteq I
\\
r_{k} = \varphi(i_{k}) \in \R.
\end{array}\right.
\end{array} \]

\noindent This explains why such an element $\varphi\in\Mlt(I)$ is
sometimes called a \emph{multiset} on $I$: it counts elements
$i_{k}\in I$ with multiplicity $r_{k} = \varphi(i_{k})\in\R$.

In this paper we shall use these monads $\Pow$ and $\Mlt$ exclusively
on natural numbers, as finite sets; in that case $\Mlt(n) = \R^{n}$,
as used above.

We shall not really use that $\Pow$ and $\Mlt$ are monads, except for
the following construction: each function $T\colon I \rightarrow
\Mlt(J)$ has a `Kleisli' or `linear' extension $T_{*} \colon \Mlt(I)
\rightarrow \Mlt(J)$ given by:
\begin{equation}
\label{eqn:extension}
\begin{array}{rcl}
T_{*}(\varphi)(j)
& = &
\displaystyle\sum_{i\in I}\, T(i)(j) \cdot \varphi(i).
\end{array}
\end{equation}

\noindent The transistion map $T$ in a layer $n\Rightarrow k$ is the
\emph{linear} part of the associated function $\R^{n} \rightarrow
\R^{k}$, and the activation function $\alpha$ is the \emph{non-linear}
part.  This linear role of $T$ is emphasised by using this linear
extension $T_*$.

Notice that if $T(i)(j) = 0$, then the input from node $i$ does not
contribute to the outcome. Hence this corresponds to not having a
connection $i \rightarrow j$ in the layer. When it comes to updating,
we have to distinguish between a weight being $0$ because there is no
connection --- so that it remains $0$ --- and weights that happen to
be zero at some point in time, but may become non-zero after an
update. This is done via the mask function $M$.
\end{remark}

\begin{definition}
\label{def:forward}
Let $\tuple{T, M, \alpha}$ be a layer $n\Rightarrow k$ as in
Definition~\ref{def:layer}. It gives rise to a (differentiable)
function $\Scottint{T, \alpha} \colon \R^{n} \rightarrow \R^{k}$
in the following manner.
\begin{equation}
\label{eqn:forward}
\begin{array}{rcl}
\Scottint{T, \alpha}(\vec{x})
& \coloneqq &
\vec{\alpha}\big(T_{*}(\vec{x},1)\big).
\end{array}
\end{equation}

\noindent Notice that we use notation $\vec{x}\in\R^n$ to indicate a
vector of reals $x_{i}\in\R$. Similarly, the notation $\vec{\alpha}$
is used to apply $\alpha\colon\R\rightarrow\R$ coordinate-wise to
$T_{*}(\vec{x},1)\in\R^{k}$, where $T_*$ is defined
in~\eqref{eqn:extension}. The additional input $1$ in
$T_{*}(\vec{x},1)$ is used to handle biases, as will be illustrated in
the example below.

The function $\Scottint{T,\alpha} \colon \R^{n} \rightarrow \R^{k}$
expresses (forward) state transformation. Sometimes we use alternative
notation $\gg$ for state transformation, defined as:
\[ \begin{array}{rcl}
(T,\alpha) \gg \vec{x}
& \,\coloneqq\, &
\Scottint{T, \alpha}(\vec{x}).
\end{array} \]

\noindent This notation is especially suggestive in combination with
loss transformation $\ll$, working backwards.
\end{definition}

The interpretation function $\Scottint{T,\alpha}$ performs what is often
called \emph{forward propagation}. We will refer to vectors
$\vec{x}\in\R^{n}$ as \emph{states}; they describe the numerical values
associated with $n$ nodes at a particular stage in a neural network. We
can then also say that forward propagation involves \emph{state
transformation}---a layer $n \Rightarrow k$ transforms states in
$\R^{n}$ to states in $\R^{k}$.

The following example\footnote{\label{fn:mazur}The example is taken from Matt Mazur's
  blog, at
  \url{https://mattmazur.com/2015/03/17/a-step-by-step-backpropagation-example/}.}
illustrates how the interpretation function works.

\begin{example}
\label{ex:mazur}
Consider the following neural network with two layers.
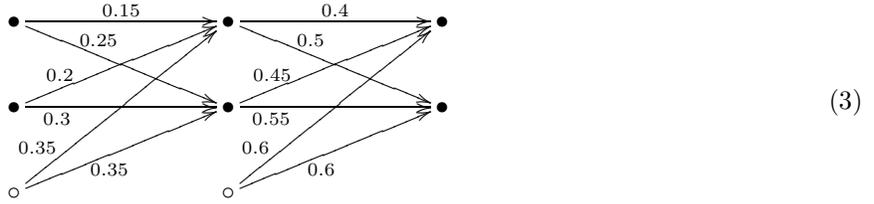
\begin{equation}
\label{nn:mazur}
\vcenter{\xymatrix@R-0.0pc@C+4pc{
\bullet\ar[r]^-{0.15}\ar[dr]^(0.35){0.25}
   & \bullet\ar[r]^-{0.4}\ar[dr]^(0.35){0.5}
   & \bullet
\\
\bullet\ar[ur]^(0.25){0.2}\ar[r]_(0.2){0.3}
   & \bullet\ar[ur]^(0.25){0.45}\ar[r]_(0.2){0.55}
   & \bullet
\\
\circ\ar[uur]^(0.2){0.35}\ar[ur]_(0.4){0.35}
   & \circ\ar[uur]^(0.2){0.6}\ar[ur]_(0.4){0.6}
}}
\end{equation}

\noindent We shall describe this network as two layers:
\[ \xymatrix@C+2pc{
2\ar@{=>}[r]^-{\tuple{T,M,\sigma}} & 
   2\ar@{=>}[r]^-{\tuple{S,M,\sigma}} & 2
}
\mbox{\qquad where \qquad}
\left\{\begin{array}{l}
M(i) = 2 = \{0,1\}
\\
\sigma(z) = \frac{1}{1+e^{-z}}
\end{array}\right. \]

\noindent In this network all connections are mutable, as indicated
via the function $M$ which sends each $i\in 2$ to the whole subset
$M(i) = 2\subseteq 2$. The activation function is the so-called
sigmoid function $\sigma$, for both layers, given by $\sigma(z) = 
\nicefrac{1}{(1+e^{-z})}$.

The two transition functions $T,S$ have type $3 \rightarrow \Mlt(2)$.
Their definition is given by the labels on the arrows in the
network~\eqref{nn:mazur}:
\[ \begin{array}{rclcrcl}
T(0)
& = &
0.15\ket{0} + 0.25\ket{1}
& \hspace*{5em} &
S(0)
& = &
0.4\ket{0} + 0.5\ket{1}
\\
T(1)
& = &
0.2\ket{0} + 0.3\ket{1}
& &
S(1)
& = &
0.45\ket{0} + 0.55\ket{1}
\\
T(2)
& = &
0.35\ket{0} + 0.35\ket{1}
& &
S(2)
& = &
0.6\ket{0} + 0.6\ket{1}.
\end{array} \]

\noindent Alternatively, one may see $T,S$ as matrices:
\[ \begin{array}{rclcrcl}
T
& = &
\left(\begin{matrix}
0.15 & 0.2 & 0.35 \\
0.25 & 0.3 & 0.35 \\
\end{matrix}\right)
& \hspace*{5em} &
S
& = &
\left(\begin{matrix}
0.4 & 0.45 & 0.6 \\
0.5 & 0.55 & 0.6 \\
\end{matrix}\right)

\end{array} \]

We thus get, according to~\eqref{eqn:forward}:
\[ \begin{array}{rcl}
\Scottint{T,\sigma}(x_{0}, x_{1})
& = &
\tuple{\,\sigma\big(T_{*}(x_{0}, x_{1}, 1)(0)\big), \; 
   \sigma\big(T_{*}(x_{0}, x_{1}, 1)(1)\big) \,}
\\
& = &
\langle\,\sigma\big(T(0)(0)\cdot x_{0} + T(1)(0)\cdot x_{1} 
   + T(2)(0)\cdot 1\big),
\\
& & \qquad 
   \sigma\big(T(0)(1)\cdot x_{0} + T(1)(1)\cdot x_{1} 
   + T(2)(1)\cdot 1\big) \,\rangle
\\
& = &
\langle\,\sigma\big(0.15\cdot x_{0} + 0.2\cdot x_{1} + 0.35\big),
   \sigma\big(0.25\cdot x_{0} + 0.3\cdot x_{1} + 0.35\big) \,\rangle
\\
\Scottint{S,\sigma}(y_{0}, y_{1})
& = &
\langle\,\sigma\big(0.4\cdot y_{0} + 0.45\cdot y_{1} + 0.6\big),
   \sigma\big(0.5\cdot y_{0} + 0.55\cdot y_{1} + 0.6\big) \,\rangle.
\end{array} \]

\noindent We see how the bias is described via the arrows out of the
`open' nodes $\circ$ in~\eqref{nn:mazur} and is added in the
appropriate manner to the outcome, via the value `$1$' on the
right-hand-side in~\eqref{eqn:forward}.

The network transforms an initial state $\tuple{0.05, 0.1}\in \R^{2}$
first into\footnote{The calculations here, and in
  Example~\ref{ex:mazurback} have been done with simple Python code,
  using the \texttt{numpy} library.} :
\[ \begin{array}{rcccl}
\Scottint{T,\sigma}(0.05, 0.1)
& = &
\tuple{\, \sigma(0.3775), \sigma(0.3925) \,}
& = &
\tuple{\, 0.59326999, 0.59688438 \,}
\end{array} \]

\noindent Subsequently it yields as final state:
\[ \begin{array}{rcccl}
\Scottint{S,\sigma}(0.59327, 0.59688)
& = &
\tuple{\, \sigma(1.10591), \sigma(1.22492) \,}  
& = &
\tuple{\, 0.75136507, 0.77292847 \,}. 
\end{array} \]

%

\end{example}

We write $\NN$ for the category of neural networks, as
in~\cite{FongST17}. Its objects are natural numbers $n\in\N$,
corresponding to $n$ nodes. A morphism $n\rightarrow k$ in $\NN$ is a
sequence of layers $n \Rightarrow \cdots \Rightarrow k$, forming a
neural network. Composition in $\NN$ is given by concatenation of
sequences; a (tagged) empty sequence is used as identity map for each
object $n$.

Next, we write $\RF$ for the category of real multivariate
differentiable functions: objects are natural numbers and morphisms
$n\rightarrow k$ are differentiable functions $\R^{n} \rightarrow
\R^{k}$.

\begin{proposition}
\label{prop:forward}
Forward state transformation (propagation) yields a functor $\NN
\rightarrow \RF$, which is the identity on objects. A morphism
$n\rightarrow k$ in $\NN$, given by a sequence of layers
$\tuple{\ell_{1}, \cdots, \ell_{m}}$, is sent to the composite
$\Scottint{\ell_{m}} \after \cdots \after \Scottint{\ell_{1}} \colon
\R^{n} \rightarrow \R^{k}$, with the understanding that an empty
sequence $\tuple{}\colon n \rightarrow n$ in $\NN$ gets sent to the
identity function $\R^{n} \rightarrow \R^{n}$. This yields a functor
by construction. \QED
\end{proposition}

In line with this description we shall interpret a morphism $N =
\tuple{\ell_{1}, \ldots, \ell_{m}} \colon n \rightarrow k$ in the
category $\NN$ as a function $\Scottint{N} = \Scottint{\ell_{m}}
\after \cdots \after \Scottint{\ell_{1}} \colon \R^{n} \rightarrow
\R^{k}$. We also write $N\gg x$ for $\Scottint{N}(x)$.

\section{Backward loss transformations}\label{sec:loss}

In the theory of neural networks one uses `loss' functions to
evaluate how much the outcome of a computation differs from a certain
`target'. A common choice is the following. Given outcomes $\vec{y}
\in \R^{k}$ and a target $\vec{t}\in\R^{k}$ one takes as loss:
\[ \textstyle\frac{1}{2}\,{\displaystyle\sum}_{i}\, (y_{i} - t_{i})^{2} \]

\noindent Here we abstract away from the precise form of such
computations and use a function $L$ for loss. In fact, we
incorporate the target $\vec{t}$ in the loss function, so that for
the above example we can give $L$ the type $L\colon \R^{k} \rightarrow
\R$, with definition:
\[ \begin{array}{rcccl}
\vec{y} \models L
& \,\coloneqq\, &
L(\vec{y})
\end{array} \]

\noindent The validity notation $\models$ emerges from the view that
vectors $\vec{y}\in\R^{k}$ are states (of type $k$), and loss
functions $L \colon \R^{k} \rightarrow \R$ are predicates (of type
$k$). The notation $\vec{y} \models L$ then expresses the value of the
loss $L$ in the state $\vec{y}$.

We now come to backward transformation of loss along a layer. We
ignore mutability because it does not play a role.

\begin{definition}
\label{def:backward}
Let $(T,\alpha) \colon n \Rightarrow k$ be a single layer. Each loss
function $L \colon \R^{k} \rightarrow \R$ on the codomain $k$ of this
layer can be transformed into a loss function $(T,\alpha) \ll L \colon
\R^{n} \rightarrow \R$ on the domain $n$ via:
\[ \begin{array}{rcl}
(T,\alpha) \ll L
& {\,\coloneqq\,} &
L \after \Scottint{T,\alpha} \;\colon\; \R^{n} \rightarrow \R^{k} \rightarrow \R.
\end{array} \]

\noindent For a morphism $N = \tuple{\ell_{1}, \ldots, \ell_{m}}
\colon n \rightarrow k$ in the category $\NN$ of neural networks we
define:
\[ \begin{array}{rcccccl}
N \ll L
& {\,\coloneqq\,} &
\ell_{1} \ll \cdots (\ell_{n} \ll E)
& = &
L \after \Scottint{\ell_{m}} \after \cdots \after \Scottint{\ell_{1}}
& = &
L \after \Scottint{N}.
\end{array} \]
\end{definition}

We can now formulate a familiar property for validity and
transformations, see \textit{e.g.}~\cite{Jacobs15,Jacobs17b}.

\begin{lemma}
\label{lem:validity}
For any neural network $N\colon n \rightarrow k$ in $\NN$, any loss
function $L: \R^{k} \to \R$ and any state $\vec{x} \in \R^{n}$, one has:
\begin{equation}
\label{eqn:validity}
\begin{array}{rcl}
N \gg \vec{x} \models L
& {\quad=\quad} &
\vec{x} \models N \ll L.
\end{array}
\end{equation}
\end{lemma}

\begin{myproof}
By the definition of these notations:
\[ \begin{array}[b]{rcl}
N \gg \vec{x} \models L
& = &
L\big(N \gg \vec{x}\big)
\\
& = &
\big(L \after \Scottint{N}\big)(\vec{x})
\\
& = &
\big(N \ll L\big)(\vec{x})
\\
& = &
\vec{x} \models N \ll L.
\end{array} \eqno{\QEDbox} \]
\end{myproof}

Many forms of state and predicate transformation can be described in
the form of a `state-and-effect triangle', where `effect' is used as
alternative name for `predicate', see~\cite{Jacobs17b}. Here this
takes the following form.

\begin{theorem}
\label{thm:triangle}
There are state and predicate functors $\Stat$ and $\Pred$ in a
triangle:
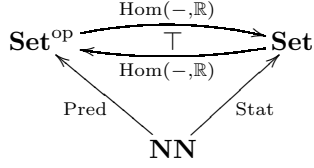
\[ \xymatrix@R+0.5pc{
\op{\Set}\ar@/^1.6ex/[rr]^{\Hom(-,\R)} & \top & \Set\ar@/^1.2ex/[ll]^{\Hom(-,\R)}
\\
& \NN\ar[ul]^{\Pred}\ar[ur]_{\Stat}
} \]

\noindent given by:
\[ \begin{array}[b]{rclclcrclcl}
\Pred(n)
& = &
\R^{\R^{n}}
& & 
& \hspace*{5em} &
\Stat(n)
& = &
\R^{n}
& &
\\
\Pred(N)
& = &
N \ll (-)
& = &
(-) \after \Scottint{N}
& & 
\Stat(N)
& = &
N \gg (-)
& = &
\Scottint{N} \after (-).
\end{array} \eqno{\QEDbox} \]
\end{theorem}

The above triangle commutes in one direction: $\Hom(-,\R) \after \Stat
= \Pred$. In order to obtain commutation in the other direction one
typically restricts the category $\Set$ to an appropriate subcategory
of algebraic structures. For instance, in probabilistic computation,
states form convex sets and predicates form effect modules, see
\textit{e.g.}~\cite{Jacobs15,Jacobs17a}. In the present situation with
neural nets it remains to be investigated which algebraic structures
are relevant. That is not so clear in the current general set up, for
instance because we impose no restrictions on the loss functions that
we use.

\section{Back propagation}\label{sec:propagation}

In the setting of neural networks, back propagation is a key step to
perform an update of (the linear part of) a layer. Here we shall give an
abstract description of such updates, in terms of a loss function $L$
as used in the previous section. In fact, we assume that what is
commonly called the learning rate $\eta$ is also incorporated in $L$.

Let $\tuple{T, M, \alpha} \colon n \rightarrow k$ be a layer. Given an
input state $\vec{a}\in\R^{n}$ and an (differentiable) loss predicate
$L\colon \R^{k} \rightarrow \R$ we will define a \emph{gradient}
\[ \gradient_{(\vec{a},L)}(T)
\mbox{\qquad and use it to change $T$ into \qquad}
T - M\odot\gradient_{(\vec{a},L)}(T), \]

\noindent where the mutability map $M\colon n \rightarrow \Pow(k)$ is
used as $k\times n$ Boolean matrix (with $0$'s and $1$'s only), and
where $\odot$ is the Hadamard product, given by elementwise
multiplication. It ensures that only mutable connections are updated.

\begin{definition}
\label{def:gradient}
In the situation just described, the gradient can be given as:
\begin{equation}
\label{eqn:gradient}
\begin{array}{rcl}
\gradient_{(\vec{a},L)}(T)
& \coloneqq &
\Big(\frac{\partial}{\partial X} 
   \big((X,\alpha) \gg \vec{a} \models L\big)\Big)(T).
\end{array}
\end{equation} 

\noindent We have introduced a new bound variable $X$, to clearly
indicate the derivative that we are interested in. The type of $X$ 
is the same as $T$, namely a $k\times(n+1)$ matrix.
\end{definition}

In order to compute this gradient, we recall that the derivative of a
(differentiable) function $f\colon \R^{n} \rightarrow \R^{m}$ is the
$m\times n$ `Jacobian' matrix of partial derivatives:
\[ \begin{array}{rcl}
f'
& = &
\left(\begin{matrix}
\frac{\partial f_{1}}{\partial x_{1}} 
& \cdots &
\frac{\partial f_{1}}{\partial x_{n}}
\\
\vdots & & \vdots
\\
\frac{\partial f_{m}}{\partial x_{1}} 
& \cdots &
\frac{\partial f_{m}}{\partial x_{n}}
\end{matrix}\right)
\end{array} \]

\begin{lemma}
\label{lem:gradient}
In the situation of Definition~\ref{def:gradient},
\begin{enumerate}
\item \label{lem:gradient:outer} The gradient
  $\gradient_{(\vec{a},L)}(T)$ can be calculated as:
\[ \begin{array}{rclcrcl}
\gradient_{(\vec{a},L)}(T)
& = &
\vec{s} \cdot (\vec{a},1)^{\textsf{\emph{T}}}
& \mbox{\qquad where \qquad} &
s_{j}
& = &
L'((T,\alpha)\gg\vec{a})_{j}\cdot \alpha'(T_{*}(\vec{a},1)_{j}).
\end{array} \]

\noindent (The superscript $\textsf{\emph{T}}$ in
$(-)^{\textsf{\emph{T}}}$ is for `matrix transpose', and is unrelated
to the transition map $T$.)

\item \label{lem:gradient:sigmoid} In the special case where $\alpha$ is
  the sigmoid function $\sigma$, the vector $\vec{s}$ in
  point~\eqref{lem:gradient:outer} is a Hadamard product:
\[ \begin{array}{rclcrcl}
\vec{s}
& = &
L'(\vec{b}) \odot \vec{b} \odot (1 - \vec{b})
& \mbox{\qquad where \qquad} &
\vec{b}
& = &
(T,\sigma) \gg \vec{a}.
\end{array} \]
\end{enumerate}
\end{lemma}

\begin{myproof}
The chain rule for multivariate functions gives a product of matrices:
\begin{equation}
\label{eqn:gradientchain}
\begin{array}{rcl}
\gradient_{(\vec{a},L)}(T)
& = &
L'\big((T,\alpha) \gg \vec{a}) \cdot \vec{\alpha}'\big(T_{*}(\vec{a},1)\big)
   \cdot \Big(\frac{\partial}{\partial X} X_{*}(\vec{a},1)\Big)(T).
\end{array}
\end{equation}

\noindent We elaborate the three parts one-by-one.
\begin{itemize}
\item The derivative of the loss function $L \colon \R^{k}
  \rightarrow \R$ is given by its partial derivatives, written as $L'
  \colon \R^{k} \rightarrow \R^{k}$. Thus, the first part
  $L'\big((T,\alpha) \gg \vec{a})$ of~\eqref{eqn:gradientchain} is in
  $\R^{k}$.

\item The derivative of the coordinate-wise application $\vec{\alpha}
  \colon \R^{k} \rightarrow \R^{k}$ of $\alpha\colon\R\rightarrow\R$,
  applied to the sequence $T_{*}(\vec{a},1)\in\R^{k}$ consists of the
  $k\times k$ diagonal matrix with entries
  $\alpha'(T_{*}(\vec{a},1)_{j})$ at position $j,j$. We shall write
  this diagonal as a vector $\tuple{\alpha'(T_{*}(\vec{a},1)_{j})} \in
  \R^{k}$.

The product of the first two factors in~\eqref{eqn:gradientchain}
can thus be written as a Hadamard (coordinatewise) product $\odot$:
\[ L'\big((T,\alpha) \gg \vec{a}) \odot \tuple{\alpha'(T_{*}(\vec{a},1)_{j})}. \]

\item For the third part in~\eqref{eqn:gradientchain} we notice that
  $X \mapsto X_{*}(\vec{a},1)$ is a function $\R^{k\times (n+1)}
  \rightarrow \R^{k}$. The $j$\textsuperscript{th} row of its Jacobian
  consists of the $k\times(n+1)$ matrix with $\vec{a},1$ at row $j$ 
  and zeros everywhere else. Indeed, the $j$\textsuperscript{th} 
  coordinate $X_{*}(\vec{a},1)_{j}$ is given by:
\[ \begin{array}{rcl}
X_{*}(\vec{a},1)_{j}
& = &
X_{j1}a_{1} + \cdots + X_{jn}a_{n} + X_{j(n+1)}.
\end{array} \]


\noindent Taking its derivative with respect to the variables $X_{ji}$
yields the $k\times(n+1)$ matrix:
\[ \left(\begin{matrix}
\vec{0} & \cdots & \vec{0} & \vec{0}
\\
a_{1} & \cdots & a_{n} & 1\rlap{\hspace*{2em}$\leftarrow \mbox{ row } j$}
\\
\vec{0} & \cdots & \vec{0} & \vec{0}
\end{matrix}\right) \]
\end{itemize}

\noindent Thus, $\big(\frac{\partial}{\partial X} X_{*}(\vec{a},1)\big)(T)$
consists of $k$-many of such matrices stacked on top of each other.

\begin{enumerate}
\item Writing $s_{j} = L'((T,\alpha)\gg\vec{a})_{j}\cdot
  \alpha'(T_{*}(\vec{a},1)_{j})$ we can put the previous three bullets
  together and write the gradient $\gradient_{(\vec{a},L)}(T)$ as an
  outer product:
\[ \begin{array}{rcccl}
\left(\begin{matrix}
a_{1}s_{1}
& \cdots &
a_{n}s_{1}
&
s_{1}
\\
\vdots & & \vdots 
\\
a_{1}s_{k}
& \cdots &
a_{n}s_{k}
&
s_{k}
\end{matrix}\right)
& = &
\left(\begin{matrix}
s_{1}
\\
\vdots
\\
s_{k}
\end{matrix}\right)
\cdot 
\left(\begin{matrix}
a_{1} & \cdots & a_{n} & 1
\end{matrix}\right)
& = &
\vec{s} \cdot (\vec{a},1)^{\textsf{T}}.
\end{array} \]

\item Directly from~\eqref{lem:gradient:outer} since $\sigma' =
  \sigma(1-\sigma)$. \QED
\end{enumerate}
\end{myproof}

Next we are interested in gradients of multiple layers.

\begin{proposition}
\label{prop:multiplegradient}
Consider two consecutive layers $m
  \smash{\stackrel{(S,\beta)}{\Longrightarrow}} n
  \smash{\stackrel{(T,\alpha)}{\Longrightarrow}} k$, with initial
  state $\vec{a}\in\R^{m}$ and loss function $L \colon \R^{k}
  \rightarrow \R$. The gradient for updating $S$ is:
\[ \begin{array}{rcl}
\Big(\frac{\partial}{\partial X} 
   \big((T,\alpha) \gg (X,\beta) \gg \vec{a} \models L\big)\Big)(S)
& \smash{\stackrel{\eqref{eqn:validity}}{=}} &
\Big(\frac{\partial}{\partial X} 
   \big((X,\beta) \gg \vec{a} \models (T,\alpha) \ll L\big)\Big)(S)
\\[+.5em]
& = &
\gradient_{(a, (T,\alpha) \ll L)}(S).
\end{array} \]

\noindent The derivative of the transformed loss $(T,\alpha) \ll L$
is by the chain rule:
\begin{equation}
\label{eqn:errortransformderivative}
\begin{array}{rcl}
\big((T,\alpha) \ll L\big)'(\vec{y})
& = &
\Big(L'\big(\Scottint{T,\alpha}(\vec{y})\big)\odot 
  \vec{\alpha}'\big(T_{*}(\vec{y},1)\big)\Big) \cdot [T],
\end{array}
\end{equation}

\noindent where $[T]$ is the $k\times n$ matrix obtained from the
$k\times (n+1)$ matrix $T$ by omitting the last column.

More generally, for appropriately typed neural nets $N,M$,
\[ \begin{array}{rcl}
\Big(\frac{\partial}{\partial X} 
   \big(N \gg (X,\alpha) \gg M \gg \vec{a} \models L\big)\Big)(S)
& = &
\gradient_{(M \gg a, N \ll L)}(S).
\end{array} \]
\end{proposition}

\begin{myproof}
The first equation in the above proposition obviously holds. We
concentrate on the second
equation~\eqref{eqn:errortransformderivative}:
\[ \begin{array}{rcl}
\big((T,\alpha) \ll L\big)'(\vec{y})
& = &
\big(L \after \Scottint{T,\alpha}\big)'(\vec{y})
\\
& = &
L'\big(\Scottint{T,\alpha}(\vec{y})\big)\cdot \Scottint{T,\alpha}'(\vec{y})
\\
& = &
\Big(L'\big(\Scottint{T,\alpha}(\vec{y})\big)\odot 
  \vec{\alpha}'\big(T_{*}(\vec{y},1)\big)\Big) \cdot T_{*}'(\vec{y},1)
\\
& = &
\Big(L'\big(\Scottint{T,\alpha}(\vec{y})\big)\odot 
  \vec{\alpha}'\big(T_{*}(\vec{y},1)\big)\Big) \cdot [T].
\end{array} \]

\noindent We still need to prove $T_{*}'(\vec{y},1) = [T]$, where
$[T]$ is obtained from $T$ by dropping the last column. The function
$T_{*}(-,1)$ has type $\R^{n} \rightarrow \R^{k}$, so the derivative
$T_{*}'(\vec{y},1)$ is a $k\times n$ matrix with entry at $i,j$ given
by:
\[ \begin{array}{rcccl}
\frac{\partial T_{*}(\vec{y},1)_{i}}{\partial y_{j}}
& = &
\frac{\partial(T_{i1}y_{1} + \cdots + T_{in}y_{n} + T_{i(n+1)})}{\partial y_{j}}
& = &
T_{ij}.
\end{array} \]

\noindent Together these $T_{ij}$, for $1\leq i\leq k$ and $1\leq j
\leq n$, form the $k\times n$ matrix $[T]$. \QED
\end{myproof}

\begin{remark}
\label{rem:losstransform}
Equation~\eqref{eqn:errortransformderivative} reveals an important
point: for actual computation of backpropagation we are not so much
interested in \emph{loss} transformation, but in \emph{erosion}
transformation, where we introduce the word `erosion' as name for the
derivative $L'$ of the loss function $L$.

For this erosion transformation we introduce new notation $\lll$. Let
$(T,\alpha) \colon n \Rightarrow k$ be a single layer, and let $E
\colon \R^{k} \rightarrow \R^{k}$ be a `erosion' function. We transform
it into another erosion function $(T,\alpha) \lll E \colon \R^{n}
\rightarrow \R^{n}$, by
following~\eqref{eqn:errortransformderivative}:
\begin{equation}
\label{eqn:losstransformation}
\begin{array}{rcl}
\big((T,\alpha) \lll E\big)(\vec{x})
& \coloneqq &
\Big(E\big(\Scottint{T,\alpha}(\vec{x})\big)\odot 
  \vec{\alpha}'\big(T_{*}(\vec{x},1)\big)\Big) \cdot [T].
\end{array}
\end{equation}

\noindent By construction we have:
\begin{equation}
\label{eqn:errorlosstransformation}
\begin{array}{rcl}
\big((T,\alpha) \ll L\big)'
& = &
(T,\alpha) \lll L'.
\end{array}
\end{equation}

\noindent Conceptually, we consider loss transformation more
fundamental than erosion tranformation, because loss transformation
gives rise to the `triangle' situation in Theorem~\ref{thm:triangle}.
In addition, erosion transformation can be expressed via derivatives
and loss transformation, as the above
equation~\eqref{eqn:errorlosstransformation} shows.

In the obvious way we can extend $\lll$
in~\eqref{eqn:losstransformation} from single to multiple layers
(neural networks). In case $\alpha$ is the sigmoid function $\sigma$,
the right-hand-side of~\eqref{eqn:losstransformation} simplifies to:
\begin{equation}
\label{eqn:sigmoidlosstransformation}
\begin{array}{rclcrcl}
\big((T,\sigma) \lll E\big)(\vec{x})
& = &
\Big(E(\vec{y}) \odot \vec{y} \odot (1-\vec{y})\Big) \cdot [T]
& \mbox{\qquad where \qquad} &
\vec{y}
& = &
(T,\sigma) \gg \vec{x}.
\end{array}
\end{equation}

\end{remark}

We illustrate back propagation for the earlier example.

\begin{example} 
\label{ex:mazurback}
We continue Example~\ref{ex:mazur} and compute the relevant gradients
for updating the transition maps/matrices $T,S$ in the
neural network~\eqref{nn:mazur} with two layers:
\[ \xymatrix@C+1pc{
2 \ar@{=>}[r]^-{(T,\sigma)} & 2 \ar@{=>}[r]^-{(S,\sigma)} & 2
} \]

\noindent We shall write input, intermediary, and final states, as
computed in Example~\ref{ex:mazur}, respectively as:
\[\begin{array}{rcl}
\vec{a}
& = &
\tuple{0.05, 0.1}
\\
\vec{b}
& = &
(T,\sigma) \gg \vec{a}
\hspace*{\arraycolsep}=\hspace*{\arraycolsep}
\tuple{0.59326999, 0.59688438}
\\
\vec{c}
& = &
(S,\sigma) \gg \vec{b}
\hspace*{\arraycolsep}=\hspace*{\arraycolsep}
\tuple{0.75136507, 0.77292847}.
\end{array} \]

\noindent The target in this example is $\tuple{0.01, 0.99} \in \R^{2}$,
so that the loss function $L \colon \R^{2} \rightarrow \R$ and its
`erosion' derivative $E = L' \colon \R^{2} \rightarrow \R^{2}$ are:
\[ \begin{array}{rclcrcl}
L(\vec{x})
& = &
\frac{1}{2}\eta\big((x_{1} - 0.01)^{2} + (x_{2} - 0.99)^{2}\big)
& {\qquad} &
E(\vec{x})
& = &
\eta\tuple{x_{1} - 0.01, x_{2} - 0.99}.
\end{array} \]

\noindent The learning rate $\eta$ is set to $0.5$.

The updating of the transition matrices $T,S$ works in backward
direction.  By Lemma~\ref{lem:gradient} we get as gradient:
\[ \begin{array}{rcl}
\gradient_{((T,\sigma) \gg \vec{a}, L)}(S)
& = &
\vec{s} \cdot (\vec{b},1)^{\textsf{T}} \mbox{\qquad where } 
   \vec{s} = E(\vec{c}) \odot \vec{c} \odot (1-\vec{c})
\\
& = &
\left(\begin{matrix}
0.08216704 & 0.08266763 & 0.13849856
\\
-0.02260254 & -0.02274024 & -0.03809824
\end{matrix}\right).
\end{array} \]

\noindent Hence the updated last transition function / matrix $S$ is:
\[ \begin{array}{rcl}
S - \vec{s} \cdot (\vec{b},1)^{\textsf{T}}
& = &
\left(\begin{matrix}
0.35891648 & 0.40866619 & 0.53075072
\\
0.51130127 & 0.56137012 & 0.61904912
\end{matrix}\right)
\end{array} \]

Our next aim is to update the preceding, first transition function /
matrix $T$.
\[ \begin{array}{rcl}
\gradient_{(\vec{a}, (S,\sigma) \ll L)}(T)
& = &
\vec{t} \cdot (\vec{a},1)^{\textsf{T}} 
\\
& & \mbox{\quad where \quad}
\begin{array}[t]{rcl}
\vec{t}
& = &
((T,\sigma) \lll E)(\vec{b}) \odot \vec{b} \odot (1-\vec{b})
\\
& \smash{\stackrel{\eqref{eqn:sigmoidlosstransformation}}{=}} &
\Big(\big(E(\vec{c}) \odot \vec{c} \odot (1-\vec{c})\big)\cdot [T]\Big)
   \odot \vec{b} \odot (1-\vec{b})
\\
& = &
\big(\vec{s} \cdot [T]\big) \odot \vec{b} \odot (1-\vec{b})
\end{array}
\\
& = &
\left(\begin{matrix}
0.00043857 & 0.00087714 & 0.00877135
\\
0.00049771 & 0.00099543 & 0.00995425
\end{matrix}\right).
\end{array} \]

\noindent The updated first matrix of the neural network is then:
\[ \begin{array}{rcl}
T - \vec{t} \cdot (\vec{c},1)^{\textsf{T}}
& = &
\left(\begin{matrix}
0.14978072 & 0.19956143 & 0.34561432
\\
0.24975114 & 0.29950229 & 0.34502287
\end{matrix}\right)
\end{array} \]

\noindent This corresponds to the numbers given in Mazur's blog
mentioned in footnote~\ref{fn:mazur}, except that there the biases are
not updated. This example illustrates that backpropagation can be done
in a recursive manner, since the values $\vec{s}$ in the first step
are re-used in $\vec{t}$ in the second step.
\end{example}

\section{Functoriality of backpropagation}

In a recent paper~\cite{FongST17} a categorical analysis of neural
networks is given.  Its main result is compositionality of
backpropagation, via a description of backpropagation as a functor. In
this section we first give a description of the functoriality of
backpropagation in the current framework, and then give a comparison
with~\cite{FongST17}.

We write $\SL$ for the category of `states and losses'.
\begin{itemize}
\item The objects of $\SL$ are triples $(n, \vec{a}, L)$, where
  $\vec{a}\in\R^{n}$ is a state of type $n$ and $L\colon \R^{n}
  \rightarrow \R$ is a (differentiable) loss function of the same
  type $n$.

\item A morphism $N\colon (n,\vec{a},L) \rightarrow (k,\vec{b},K)$ is
  a neural network $N\colon n \rightarrow k$, in the category $\NN$,
  such that both: $\vec{b} = N \gg \vec{a}$ and $K = N\ll L$.
\end{itemize}

\noindent There is an obvious forgetful functor $\mathcal{U} \colon
\SL \rightarrow \NN$ given by $\mathcal{U}(n,\vec{a},L) = n$ and
$\mathcal{U}(N) = N$.

\begin{definition}
\label{def:backpropfun}
Define \emph{backprop} $\mathcal{B} \colon \SL \rightarrow \NN$ in
the following way. On objects, we simply take
$\mathcal{G}(n,\vec{a},L) = n$. Next, let $N = \tuple{\ell_{1},
  \ldots, \ell_{m}}$ be a morphism $(n,\vec{a},L) \rightarrow
(k,\vec{b},K)$ in $\SL$, where $\ell_{i} = (T_{i}, \alpha_{i},
M_{i})$. We write:
\begin{itemize}
\item $\vec{a}_{0} \coloneqq \vec{a}$ and $\vec{a}_{i+1} \coloneqq
  \ell_{i} \gg \vec{a}_{i}$; this gives a list of states
  $\tuple{\vec{a}_{0}, \vec{a}_{1}, \ldots, \vec{a}_{m}}$ with
  $\vec{a}_{m} = \vec{b}$, by assumption;

\item $K_{m} \coloneqq K$ and $K_{i-1} \coloneqq \ell_{i} \ll K_{i}$;
  this gives a list of loss functions $\tuple{K_{0}, \ldots, K_{m}}$
  with $K_{0} = L$.
\end{itemize}

\noindent Then $\mathcal{B}(N) \colon n \rightarrow k$ is defined as a
list of layers, of the same length $m$ as $N$, with components:
\[ \begin{array}{rcl}
\mathcal{B}(N)_{i}
& \coloneqq &
\tuple{T_{i} - M_{i} \odot \gradient_{(\vec{a}_{i-1}, K_{i})}(T_{i}), \,
   \alpha_{i}, \, M_{i}}.
\end{array} \]

\noindent (Recall, $M_i$ is a Boolean `mask' matrix that takes care of
mutability, and $\odot$ is the Hadamard product.)
\end{definition}

\begin{theorem}
\label{thm:backpropfun}
Backprop $\mathcal{B} \colon \SL \rightarrow \NN$ is a functor.
\end{theorem}

\begin{myproof}
This is `immediate', but writing out the details involves a bit of
book keeping.  Let $\smash{(n,\vec{a},L) \stackrel{N}{\longrightarrow}
  (k,\vec{b},K) \stackrel{P}{\longrightarrow} (l,\vec{c}, F)}$ be
(composible) morphisms in $\SL$, where $N = \tuple{\ell_{1}, \ldots,
  \ell_{u}}$ and $P = \tuple{p_{1}, \ldots, p_{v}}$. We write
$\ell_{i} = \tuple{T^{N}_{i}, \alpha^{N}_{i}, M^{N}_{i}}$ and
similarly $p_{j} = \tuple{T^{P}_{j}, \alpha^{P}_{j}, M^{P}_{j}}$. The
procedures in the two bullets in Definition~\ref{def:backpropfun}
yield for the maps $N$ and $K$ separately:
\begin{itemize}
\item $\tuple{\vec{a}_{0}, \ldots, \vec{a}_{u}}$ and
  $\tuple{\vec{b}_{0}, \ldots, \vec{b}_{v}}$ where $\vec{a}_{0} =
  \vec{a}, \vec{a}_{i+1} = \ell_{i} \gg \vec{a}_{i}$ and $\vec{b}_{0}
  = \vec{b}, \vec{b}_{j+1} = p_{j} \gg \vec{b}_{i}$; we have
  $\vec{b}_{0} = \vec{b} = N\gg \vec{a} = N \gg \vec{a}_{0} =
  \vec{a}_{u}$;

\item $\tuple{K_{0}, \ldots, K_{u}}$ and $\tuple{F_{0}, \ldots,
  F_{v}}$ with $K_{u} = K, K_{i-1} = \ell_{i} \gg K_{i}$ and $F_{v} =
  F, F_{j-1} = p_{j} \gg F_{j}$; then $K_{u} = D = P \ll F = P \ll
  F_{v} = F_{0}$.
\end{itemize}

\noindent From the perspective of the composite sequence
$\tuple{\ell_{1}, \ldots, \ell_{u}, p_{1}, \ldots, p_{v}}$ we can go
through the same process and obtain sequences $\tuple{\vec{a}'_{0},
  \ldots, \vec{a}'_{u+v}}$ and $\tuple{F'_{0}, \ldots, F'_{u+v}}$
with:
\[ \begin{array}{rclcrcl}
\vec{a}'_{i}
& = &
\left\{\begin{array}{ll}
\vec{a}_{i} \quad & \mbox{if } i \leq p
\\
\vec{b}_{i-p} & \mbox{otherwise}
\end{array}\right.
& \mbox{\qquad} &
F'_{j}
& = &
\left\{\begin{array}{ll}
K_{j} \quad & \mbox{if } j \leq p
\\
F_{j-p} & \mbox{otherwise.}
\end{array}\right.
\end{array} \]

\noindent We can now describe the components of the updated network
$\mathcal{B}(P \after N)$. For $1\leq i\leq u$ and $1\leq j\leq v$,
\[ \begin{array}[b]{rcl}
\mathcal{B}(P \after N)_{i}
& = &
\tuple{T^{N}_{i} - M^{N}_{i} \odot \gradient_{(\vec{a}'_{i-1}, F'_{i})}(T^{N}_{i}), \,
   \alpha^{N}_{i}, \, M^{N}_{i}}
\\
& = &
\tuple{T^{N}_{i} - M^{N}_{i} \odot \gradient_{(\vec{a}_{i-1}, K_{i})}(T^{N}_{i}), \,
   \alpha^{N}_{i}, \, M^{N}_{i}}
\\
& = &
\mathcal{B}(N)_{i}
\\
& = &
\big(\mathcal{B}(P) \after \mathcal{B}(N)\big)_{i}
\\
\mathcal{B}(P \after N)_{u+j}
& = &
\tuple{T^{P}_{j} - M^{P}_{j} \odot \gradient_{(\vec{a}'_{u+j-1}, F'_{u+j})}(T^{P}_{j}), \,
   \alpha^{P}_{j}, \, M^{P}_{j}}
\\
& = &
\tuple{T^{P}_{j} - M^{P}_{j} \odot \gradient_{(\vec{b}_{j-1}, F_{j})}(T^{P}_{j}), \,
   \alpha^{P}_{j}, \, M^{P}_{j}}
\\
& = &
\mathcal{B}(P)_{j}
\\
& = &
\big(\mathcal{B}(K) \after \mathcal{B}(N)\big)_{p+j}.
\end{array} \eqno{\QEDbox} \] 
\end{myproof}

We conclude this section with a comparison to~\cite{FongST17}, where
it was first shown that backpropagation is functorial.  The approach
in~\cite{FongST17} is both more abstract and more concrete than
ours.
\begin{enumerate}
\item Here, a layer $(T,\alpha) \colon n\Rightarrow k$ of a neural
  network consists of linear part $T\colon n+1\rightarrow \Mlt(k)$ and
  a non-linear part $\alpha\colon \R \rightarrow \R$. We ignore the
  mutability matrix $M$ for a moment.  As shown in
  Definition~\ref{def:forward}, the layer $(T,\alpha)$ gives rise to
  an interpretation function $\Scottint{T,\alpha} \colon \R^{n}
  \rightarrow \R^{k}$ that performs forward state transformation
  $(T,\alpha) \gg (-)$. In~\cite{FongST17} there is no such concrete
  description of a layer. Instead, the paper works with `parametrised'
  functions $P\times \R^{n} \rightarrow \R^{k}$. Our approach fits in
  this framework by taking the set of linear parts $P = \Mlt(k)^{n+1}$
  as parameter set. These parametrised functions are organised in a
  category $\textsf{Para}$, which is shown to be symmetric monoidal
  closed.

\item The comparison of the outcome of a state transformation by a
  network $n\Rightarrow k$ and a target $\vec{t}\in \R^{k}$ is
  captured here abstractly via a loss function $L \colon \R^{k}
  \rightarrow \R$. This more general perspective allows us to define
  loss transformation $N \ll L$ along a network $N$. We have thus
  developed a view on neural network computation, with forward and
  backward transformations, that is in line with standard approached
  to (categorical) program semantics. It gives rise to the pattern of
  a state-and-effect triangle in Theorem~\ref{thm:triangle}. Moreover,
  we show that there is an associated `erosion transformation'
  function, that is suitable related to loss transformation via
  derivatives, see~\eqref{eqn:errorlosstransformation}.

In the formalism of~\cite{FongST17} backward computation also plays a
role, via a function `$r$', of type $P \times \R^{n} \times \R^{k}
\rightarrow \R^{n}$, for a network $n\Rightarrow k$. It corresponds to
our erosion transformation~\eqref{eqn:losstransformation}, roughly as:
$r(\ell, \vec{a}, \vec{b}) = (\ell \lll L'_{\vec{b}})(\vec{a})$, where
$L'_{\vec{b}}$ is the derivative of the loss function $L_{\vec{b}}$
associated with the `target' $\vec{b}$.

\item Here we have concentrated on the sequential structure.
  In~\cite{FongST17}, parallel composition is also taken into account
  in the form of symmetric monoidal structure. For us, such additional
  structure is left as future work.
\end{enumerate}

\section{Conclusions}\label{sec:conclusions}

In this paper, we have examined neural networks as programs in a
state-and-effect framework. In particular, we have characterized the
application of a neural network to an input as a kind of state
transformation and backpropagation of loss along the network as a kind
of predicate transformation on losses. We also observed that the
compositionality of backpropagation corresponds to the functoriality
of a mapping between a category of states-and-effects to the category
of neural networks.

For the sake of illustrating this perspective on neural networks, we
have deliberately chosen a simple subclass of the known network
architectures and built a category of multilayer perceptron
(MLPs). However, we believe it is possible to develop a richer
categorical structure capable of capturing a much wider variety of
network architectures. This may be the focus of future work.

We also considered a single training scheme: backpropagation paired
with stochastic gradient descent (with a fixed learning rate). We are
interested in modeling other kinds of neural network training
categorically.

As mentioned in the discussion following Theorem~\ref{thm:triangle},
there is typically a category of algebraic structures in the upper
right vertex of the state-and-effect triangle which we have not
determined yet.

\subsection*{Acknowledgments} The first author (BJ) acknowledges
support from the European Research Council under the European Union's
Seventh Framework Programme (FP7/2007-2013) / ERC grant agreement
n\textsuperscript{o} 320571. The second author (DS) is supported by
the JST ERATO HASUO Metamathematics for Systems Design Project
(No.\ JPMJER1603).

\bibliographystyle{entcs}


\end{document}